\documentclass{article}
\usepackage{graphicx} 
\usepackage{caption}
\usepackage{subcaption}

\usepackage[final, nonatbib]{neurips_2019}

\usepackage[utf8]{inputenc} 
\usepackage[T1]{fontenc}    
\usepackage{hyperref}       
\usepackage{url}            
\usepackage{booktabs}       
\usepackage{amsfonts}       
\usepackage{nicefrac}       
\usepackage{microtype}      

\usepackage{todonotes}

\title{Synthetic Event Time Series Health Data Generation}

\author{
Saloni Dash \\
BITS Pilani, Goa Campus \\
Goa, India \\
\texttt{salonidash77@gmail.com} 
\And
Ritik Dutta \\
IIT Gandhinagar \\
Gandhinagar, India \\
\texttt{dutta.ritik@iitgn.ac.in}
\And
Isabelle Guyon \\
UPSud/INRIA U. Paris-Saclay \\
Paris-Saclay, France \\
\texttt{guyon@clopinet.com}
\And
Adrien Pavao \\
UPSud/INRIA U. Paris-Saclay \\
Paris-Saclay, France \\
\texttt{adrien.pavao@gmail.com}
\And
Andrew Yale \\
Rensselaer Polytechnic Institute \\
Troy, New York \\
\texttt{yalea@rpi.edu} \\
\And
Kristin P. Bennett \\
Rensselaer Polytechnic Institute \\
Troy, New York \\
\texttt{bennek@rpi.edu}
}

\begin{document}

\maketitle

\begin{abstract}
Synthetic medical data which preserves privacy while maintaining utility can be used as an alternative to real medical data, which has privacy costs and resource constraints associated with it. At present, most models focus on generating cross-sectional health data which is not necessarily representative of real data. In reality, medical data is longitudinal in nature, with a single patient having multiple health events, non-uniformly distributed throughout their lifetime. These events are influenced by patient covariates such as comorbidities, age group, gender etc. as well as external temporal effects (e.g. flu season). While there exist seminal methods to model time series data, 
it becomes increasingly challenging to extend these methods to medical event time series data. Due to the complexity of the real data, in which each patient visit is an event, we transform the data by using summary statistics to characterize the events for a fixed set of time intervals,  to facilitate analysis and interpretability. We then train a generative adversarial network to generate synthetic data.  We demonstrate this approach by generating human sleep patterns, from a publicly available dataset. We empirically evaluate the generated data and show close univariate resemblance between synthetic and real data. However, we also demonstrate how stratification by covariates is required to gain a deeper understanding of synthetic data quality.   

\end{abstract}

\section{Introduction}
 Advancements in health informatics have largely been aided by the use of massive amounts of medical data. Researchers build computational models with this data, which facilitate improved healthcare provisions for the general public. Examples include drug discovery, personalized medicine and medical education. 
 However, most of the Electronic Health Records (EHR) accumulated by health organizations are not easily accessible to the public due to privacy concerns.  This stifles healthcare research. EHR datasets are rarely released, making research unreproducible. 
 Research is biased towards the few publicly available de-identified medical datasets which address only limited aspects of health care, for example the MIMIC-III (Medical Information Mart for Intensive Care) dataset. \cite{johnson2016mimic}.

Generating synthetic health data which maintains the utility of the data as well as preserves the privacy of the patients is a potential solution. Machine Learning models like Bayesian Networks \cite{avino2018generating}, Generative Adversarial Networks \cite{medgan,yale2019privacy} (medGAN, HealthGAN) etc. have been used successfully to generate health data. Nonetheless, the data generated by these methods is not representative of real medical records, as they contain only one record per patient. In real life, patient data consists of stream of in-patient and out-patient visits and other treatment events through time. Due to it's temporal nature, longitudinal data is conducive for causal analysis studies and in overall offers greater utility when compared to cross-sectional data. In addition, covariates such as age, gender, and race are critical for retrospective observational studies. Therefore, building models for generating longitudinal synthetic data with such covariates could greatly facilitate healthcare research. 




Our ultimate goal is to build an end-to-end system that generates longitudinal synthetic health data which captures temporal relations of patient records along with their covariates, preserves utility as well as privacy, and is computationally efficient, scalable and free of cost. This system should ideally not require human domain knowledge and should provide quantitative assessments of privacy, resemblance, and utility instead of just relying on theoretical guarantees. There exist few methods that generate time series medical data. Synthea \cite{synthea} uses modules informed by clinicians and real-world statistics to simulate patient records from birth to present day. It guarantees privacy as the records are not generated using real patients and claims to preserve utility by using health care practitioners and real statistics to build rules that synthesize this data.  But the time series of events that it produces do not necessarily closely model actual patient trajectories especially for patients with multiple comorbidities.
Methods such as the Recurrent Conditional GAN (RCGAN) \cite{esteban2017real}  successfully synthesize time series of continuously-captured real-valued vitals (e.g. heart rate, respiratory rate, mean arterial pressure etc.)  But how this model can be extended to generate events that happen at irregular intervals with covariates requires further research. 

\section{Method}
We demonstrate our approach by generating sleep patterns using a publicly available health dataset. We use as inspiration a sleep study that examined habitual sleep times with respect to covariates including gender, age, and day of the week \cite{basner2007american}.  The event data comes from American Time Use Survey (ATUS), a federally administered, continuous survey on time use in the United States sponsored by the Bureau of Labor Statistics and conducted by the U.S. Census Bureau \cite{atus} 
The survey measures how people divide their time among life’s activities in a nationally representative sample. There are many different types of events per person.  However, we choose to restrict our data to only the sleep activities of the people for primarily two reasons. Firstly, sleeping patterns are intuitive to understand, analyse and are extremely sensitive to covariates such as age or day of the week. Secondly, both short and long habitual sleep times are closely associated with increased mortality risks \cite{hammond1964some}, diabetes \cite{gottlieb2005association} and hypertension \cite{gottlieb2006association} which makes the sleep activity data highly suitable for causal and preemptive studies, as explained previously. 

Our strategy is to transform event data consisting of one record per event, into cross-sectional data consisting of one record per subject.  For each person, the day starts at 4:00 am on the same day and ends at the hour they wake up the next day. 
ATUS also includes covariates for each patient and event. To ensure uniformity, we limit the day to 10:00 am the next day, which gives us a 30 hour day for each person. We choose 10:00 am because 96\% of the people are awake by that time. Consequently, we divide the 30 hour day into 30 events of one hour each, and compute a summary statistic of sleep in that hour, i.e. the number of minutes spent sleeping in that hour. We then treat these events as features of the dataset and append them to the covariates of age, sex, day of the week and month of the year. In this manner, we transform the data to cross-sectional data consisting of 34 (including covariates) features per patient (a matrix) while retaining it's temporal properties.

Once the data is transformed to matrix form any method can be used to generate the data.  
For our preliminary work,  we used  HealthGAN\cite{yale2019privacy}, a Wasserstein Generative Adversarial Network, to generate our synthetic dataset. Yale et. al proposed HealthGAN and provided an empirical assessment of different data generators and quantified utility and privacy losses. Out of the several baseline methods including Gaussian Multivariate \cite{duda2012pattern} and Parzen Windows \cite{parzen1962estimation}, their proposed model, the HealthGAN, outperformed all of them in terms of privacy preservation as well as maintaining data utility. We use the HealthGAN directly to generate our synthetic dataset. HealthGAN is well suited for this task becuase it is designed to perform well on binary features. The 30 average sleep features  are almost binary since they are 0 when a person awake and 60 if a person sleeps for the entire hour. We use the data transformation method detailed in the HealthGAN paper \cite{yale2019privacy} to process our data. Other generative methods \cite{xiao2017joint,esteban2017real} are left for future work.


\section{Experimental Results}

We first determine whether the generator is able to capture the average sleep patterns of the population. Figure 2 shows that the synthetic average sleep per hour closely resembles the real data in the average case. An individual is awake by 10:00 am and asleep by 12:00 am. The generator is also able to learn covariate statistics reasonably well. As shown in Figure 3, the covariate probabilities of the real and synthetic data lie closely along the diagonal. For instance, 49\%  of the days are week days (derived from the day of the week covariate) in the real data and 48\%  of the days are week days in the synthetic data. Similar probabilities are derived from the age, sex and month covariates and are compared in the probability plot in Figure 3. 

\begin{figure}[h]
\centering
\begin{minipage}{.5\textwidth}
  \centering
  \captionsetup{width=.9\linewidth}
  \includegraphics[trim={0 0 0 0.9cm},clip,  width=.9\linewidth, height =5cm]{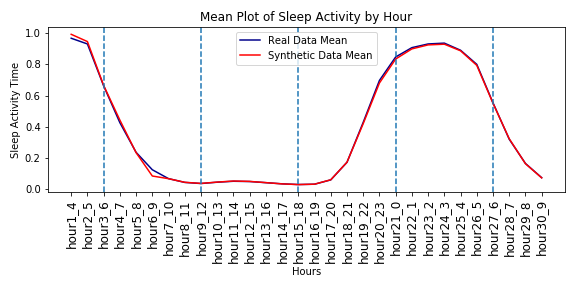}
  \captionof{figure}{\textbf{Average sleep time per hour}\\ The synthetic data mean closely matches the real data mean. hour1\_4 should be interpreted as the first hour which is 4am, hour10\_13 is the tenth hour which is 1pm and so on. The dotted blue lines indicate standard hours of the day in 6 hour intervals (6am, 12pm, etc.).}
  \label{fig:meanplot}
\end{minipage}%
\begin{minipage}{.5\textwidth}
  \centering
  \captionsetup{width=.9\linewidth}
  \includegraphics[trim={0 0 0 1.2cm},clip, width=.9\linewidth, height = 5cm]{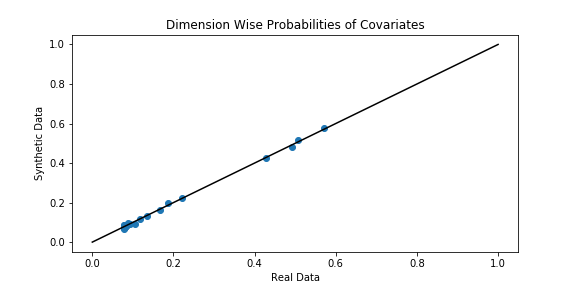}
  \captionof{figure}{\textbf{Covariate Probabilities}\\All points roughly lie along the diagonal, which shows that the covariate probabilities of the synthetic data closely match the probabilities of the real data. For every point, the x and y coordinates are covariate probabilities in the real and synthetic data respectively.}
  \label{fig:dimprob}
\end{minipage}
\end{figure}

In order to qualitatively assess the utility of the generated data, we examine the average sleep time by age and day of the week as analysed by Basner et. al \cite{basner2007american} in their sleep study.    Such analysis with respect to covariates is very typical of health care studies. 
\begin{figure}[h]
\centering
\begin{subfigure}{.5\textwidth}
  \centering
  \includegraphics[trim={0 0 0 1.4cm},clip, width=\linewidth, height =5cm]{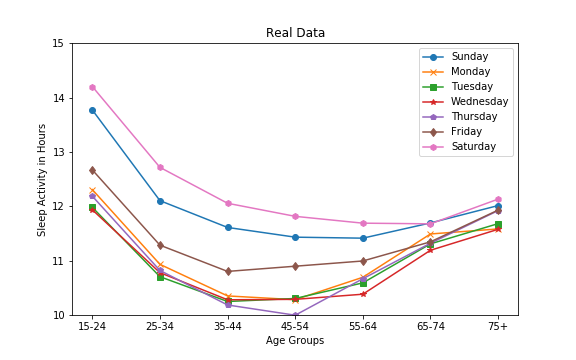}
  \caption{Real Data}
  \label{fig:meanplot}
\end{subfigure}%
\begin{subfigure}{.5\textwidth}
  \centering
  \includegraphics[trim={0 0 0 1.4cm}, clip, width=\linewidth, height = 5cm]{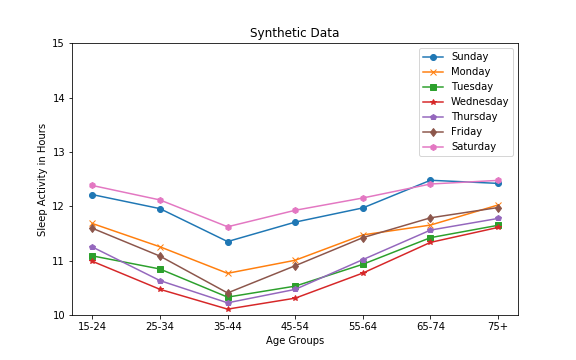}
  \caption{Synthetic Data}
  \label{fig:dimprob}
\end{subfigure}
\caption{\textbf{Average sleep time grouped by age group and day of the week}.}
\end{figure}

We observe certain distinct trends when we examine the data stratified with respect to the covariates age group and day of the week. In Figure 4(a) in the real data, the average sleep time for weekends is different than week days across all age groups. On an average, teenagers and young adults sleep (age group 15-24) significantly more than another age group, especially on weekends (Sat-Sun). Adults between 35-54 years in general require less sleep than other age groups, and with older age groups, the average sleep time increases proportionally. In Figure 4(b), the synthetic data successfully captures these trends but underestimates the average sleep time for teens in general. This may be due to irregularities and high variances in the sleep patterns of young people. 

To investigate this further, we plot the median, first and third quartiles of average sleep per hour of people aged between 15-24. Figure 5 shows the quantile plot of the average sleep times for the age group 15-24 years. It can be observed from the plot that there is high variance in the waking and sleeping times in the real data which is not captured by the synthetic data. Thus we can see
that for health data, it is not sufficient to simply report average case behavior.   It is essential to investigate behavior by covariates to make sure that patterns with respect to critical groups are well modeled.

\begin{figure}[h]
\centering
\includegraphics[width=\linewidth, height = 5cm]{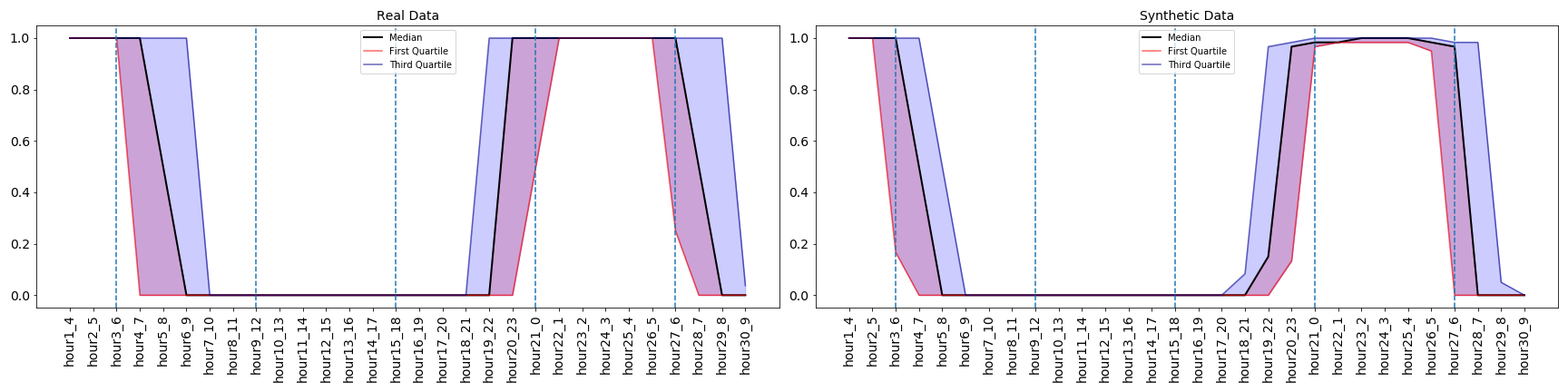}
\caption{\textbf{Quantile Plot of Age Group 15-24}\\ }
\end{figure}

\section{Conclusion and Future Work}
The potential of generating synthetic health data which respects privacy and maintains utility is groundbreaking. A significant amount of research has been conducted for generating cross-sectional data, however the problem of generating event based 
time series health data, which is illustrative of real medical data has largely been unexplored. We stress the importance and utility of generating such data with a special emphasis on the covariates; an exceedingly significant aspect of medical time series data. 

We provide an innovative methodology of transforming the longitudinal data to cross-sectional data without sacrificing the temporal properties, and then using state of the art generators to synthesize the transformed data.  The benefit of this approach is that it can be readily used with any synthetic data generation method and  facilitates effective evaluation of the quality of synthetic data. 
We test our model by synthesizing sleep patterns of Americans from a publicly available dataset and empirically show close univariate and covariate resemblances between the real and synthetic data. We also evaluate the utility of the generated data by comparing sleep trends inspired from a sleep study, where we conclude that while the generator is able to capture significant general trends but faces challenges on  subgroups with irregular sleep patterns. 
We underscore the importance of empirically evaluating the quality of synthetic medical data with respect to critical covariates.   We leave application of these methods to actual EHR data with many event types and using alternative data generators for future work.   

\bibliographystyle{plain}
\bibliography{ref}

\end{document}